%% file: anonymous-submission-latex-2026.tex
\documentclass[letterpaper]{article} 
\usepackage{aaai2026}  
\usepackage{times}  
\usepackage{helvet}  
\usepackage{courier}  
\usepackage[hyphens]{url}  
\usepackage{graphicx} 
\usepackage{natbib}  
\usepackage{caption} 
\usepackage{algorithm}
\usepackage{algorithmic}

\usepackage{newfloat}
\usepackage{listings}

\usepackage{booktabs} 
\usepackage{todonotes}
\usepackage{amsmath}
\usepackage{placeins}
\usepackage[most]{tcolorbox}
\tcbuselibrary{listingsutf8}
\usepackage{multirow}

\DeclareCaptionStyle{ruled}{labelfont=normalfont,labelsep=colon,strut=off} 
\floatstyle{ruled}
\newfloat{listing}{tb}{lst}{}
\floatname{listing}{Listing}
\title{Reinforcing VLMs to Use Tools for Detailed Visual Reasoning Under Resource Constraints}
\author{
    Sunil Kumar\textsuperscript{\rm 1}\equalcontrib,
    Bowen Zhao\textsuperscript{\rm 1}\equalcontrib,
    Leo Dirac\textsuperscript{\rm 1},
    Paulina Varshavskaya\textsuperscript{\rm 1}
}
\affiliations{
    \textsuperscript{\rm 1}Groundlight AI\\
    911 East Pike Avenue, Suite 200\\
    Seattle, WA 98122 USA\\
    sunil@groundlight.ai, bowen@groundlight.ai\\
}

\nocopyright
\begin{document}

\maketitle

\begin{abstract}
    Despite tremendous recent advances in large model reasoning ability, vision-language models (VLMs) still struggle with detailed visual reasoning, especially when compute resources are limited. To address this challenge, we draw inspiration from methods like DeepSeek-R1 for VLMs and train smaller-scale models with Group Relative Policy Optimization (GRPO) to use external tools such as zoom. The greatest benefit is obtained with a combination of GRPO learning, a simple reward structure, a simplified tool-calling interface, allocating additional tokens to the result of the tool call, and a training data mix that over-represents visually difficult examples. Compared to similarly-sized baseline models, our method achieves better performance on some visual question-answering (VQA) tasks, thanks to the detailed visual information gathered from the external tool.
\end{abstract}

\input{sections/introduction}

\input{sections/method}

\input{sections/experiments}

\input{sections/related_work}

\section{Conclusion}
We have presented a method for efficiently training small VLMs to understand visual details via GRPO-based reinforcement learning with an option to call external tools, namely to zoom into a region of interest on an image for a closer look. We have shown that this approach allows VLMs to see finer detail where it matters, thereby improving performance on some high-resolution visual understanding benchmarks compared to baselines on the same parameter scale, and even some larger engineered visual search architectures. We have also given a recipe for success involving the right data mix, reward structure, tool interface, and allocation of tokens to external tool results. 

\section*{Acknowledgments}
This research was funded by Groundlight AI.

{\small \bibliography{aaai2026}}

\appendix
\input{appendix/prompt}


\end{document}

%% file: sections/introduction.tex
\section{Introduction}

Recent advances in vision-language models (VLMs) have significantly improved performance across a spectrum of multimodal tasks. However, despite the superb capabilities of state-of-the-art models, VLMs suffer from notable limitations in processing fine-grained visual details~\citep{Rahmanzadehgervi_2024_ACCV}. Recent studies highlight that VLMs often struggle with tasks requiring detailed visual understanding, where the subtle visual details that humans can easily interpret are neglected by VLMs~\citep{tong_eyes_2024}. This issue is further amplified with high-resolution image inputs~\citep{Wu_2024_CVPR,Wang_Ding_Zeng_Zhou_Shen_Luo_Yu_Tao_2025}.

To overcome this challenge, previous research has explored the incorporation of visual search and multi-resolution processing techniques into VLM-based systems. Although visual search systems provide additional information on keypoints in images in the VLM, they can suffer from error propagation due to their complex system design~\citep{Wu_2024_CVPR,shen2024zoomeye}. Multi-resolution processing methods, on the other hand, add extra visual tokens which represent the crops of the original input image at different resolutions~\citep{thapa2024dragonfly}. While those methods may confer a performance boost, they also introduce computational overhead due to additional image tokens. Additionally, the multi-resolution image tokens themselves still represent general information about the image, and not necessarily the pertinent details of the specific region of interest (ROI) in the image that would be most helpful for the task.

In this paper, we take inspiration from recent demonstrations of tool use in large language models (LLMs) via reinforcement learning (RL) and extend this approach to multimodal VLMs with a particular focus on efficient use of limited resources. In particular, we efficiently train smaller VLMs via GRPO to capture visual details from task-specific ROIs in real-world images by appropriately calling a zoom-in tool. We choose a RL approach also due to a lack of existing data with annotated tool-use trajectories that could act as ground truth for supervised fine-tuning of a vision model for detailed visual reasoning. 

Our contributions are as follows:
\begin{itemize}
    \item We propose a recipe for extending GRPO to tool use in the visual domain with efficiency under constrained resources.
    \item We establish a set of parameters that enable a small VLM to learn tool use for visual detail understanding under resource constraints; and we find that the structure of the reward function and the data mix significantly impact success. 
    \item We present experimental results showing improvement on small-model SOTA on high-resolution VQA datasets.
    \item We discuss how the choice of external tool interface, including tokens allocated to the representation of the results of external tool calls, affects performance. 
\end{itemize}

Our findings demonstrate that even small VLMs can learn to invoke tools effectively to capture fine-grained visual details without access to ground-truth tool use trajectories or extensive post-training compute. This work provides an efficient framework for equipping VLMs with decision-making capabilities to handle spatial reasoning tasks.

%% file: sections/method.tex
\section{Method}
\begin{figure*}
    \centering
    \includegraphics[width=0.95\linewidth]{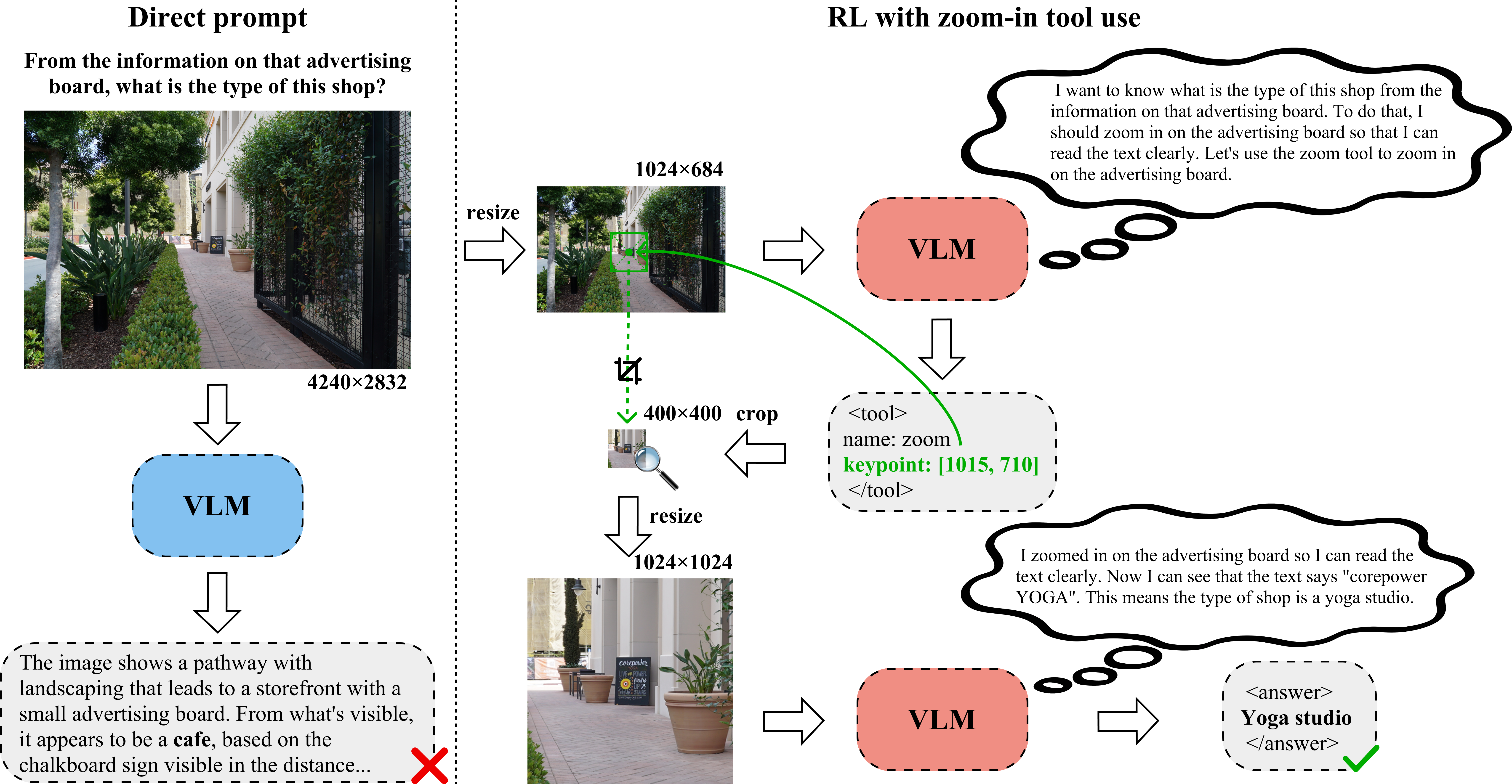}
    \caption{In contrast to direct prompting where the model fails to capture the visual details in high-resolution images, RL incentivizes the VLM to use a zoom-in tool to get extra information from the specific ROI to answer vision-oriented questions correctly (bottom). Resizing the image makes the training efficient while preserving the visual details to be noticeable to the VLM.}
    \label{fig:enter-label}
\end{figure*}

Our approach is to teach a VLM to use external tools to enhance detailed image understanding via an agentic take on reinforcement learning. In particular, we train a small Qwen2.5-VL-3B-Instruct\footnote[1]{All references to the Qwen2.5-VL models in this paper, whether ours or baselines, refer to the instruction-tuned version of these models.} model with Group Relative Policy Optimization (GRPO)~\citep{guo2025deepseek,shao2024deepseekmath} to use a zoom tool in order to find relevant information in a curated set of training examples. At the time of writing, this model is among the smallest state of the art VLMs.  We operate under a limited computational and memory budget of four NVIDIA A100 80GB GPUs. Therefore, our method emphasizes efficient use of limited resources for maximum visual understanding. This necessitates limiting image resolution and the number of tool calls during training, and additionally requires an easy-to-learn tool interface. Note that our method does not rely on ground truth trajectories of tool use, since none are generally available for visual data. Instead, given the option of calling on a tool and a reward signal, the model learns how to effectively use the tool to solve the problem, which improves downstream Q\&A accuracy.

\subsection{\label{method:details}Efficient tool use under limited resources}

\paragraph{Tool implementation}
Our zoom tool provides a focused view of a region of interest in the image, enabling the model to access high-resolution visual details that may not be discernible in the input. The tool accepts a keypoint, specified as $[x,y]$ coordinates within the image, and returns a 400 x 400 pixel square crop centered on that point. If the keypoint is near the image boundary, the crop is shifted to remain within image bounds while maintaining its fixed size. We upscale the square-cropped region such that its dimension matches that of the longer dimension of the input image, while preserving the square aspect ratio. The tool's output is injected directly into the conversation as additional image tokens, appended after the model's initial tool-calling response as a message from the user. This allows the model to generate its subsequent response with access to both the original image and the zoomed in crop, potentially enabling it to reference fine-grained visual details that were previously indiscernible. The conversation flow thus becomes: (1) initial image and question, (2) model's tool-calling response, (3) injected crop from the tool, and (4) model's answer incorporating information from both images.

\paragraph{Image resolution}
In order to limit the memory footprint during training, we downsize images such that the long side does not exceed 1024 pixels. Note that when using the zoom tool, the model only has access to this resized image and not the original higher-resolution one. To make sure the model is able to capture the details in the zoomed-in image, we further upscale the zoomed-in crop to the same size as the downsized input image - at most 1024 x 1024 pixels. We find that this strategy helps the model generalize well on high-resolution visual tasks at inference time, which will be discussed in the following sections. In contrast, during evaluation we do not resize images: the zoom-in tool is directly applied to the original high-resolution image to give the model more visual detail.

\paragraph{Limited tool use}
To reduce GPU memory load, we limit the number of zoom-in tool use calls to 1 at both training and inference times. Limiting the number of tool calls prevents out-of-memory errors from multiple sequences of image tokens appended to the conversation.  This limits our approach, as implemented, to tasks that require at most one detailed ROI per image. The model’s initial response (before any tool is called) is generated freely without structural constraints, allowing it to decide whether or not to invoke a tool based on the question. Once the tool is used, we guide the decoding of the model’s second response using the regular expression \texttt{[\^{}<]*)</think>\allowbreak{}([\^{}<]*)\allowbreak{}<answer>([\^{}<]*)</answer>}. This both prevents the model from invoking the tool, and ensures that there is an $\texttt{<answer>}$ field in the response and an end to the conversation.

\paragraph{Easy-to-learn tool interface} 
\begin{figure}
\vspace{-2em}
  \begin{center}
    \includegraphics[width=0.4\textwidth]{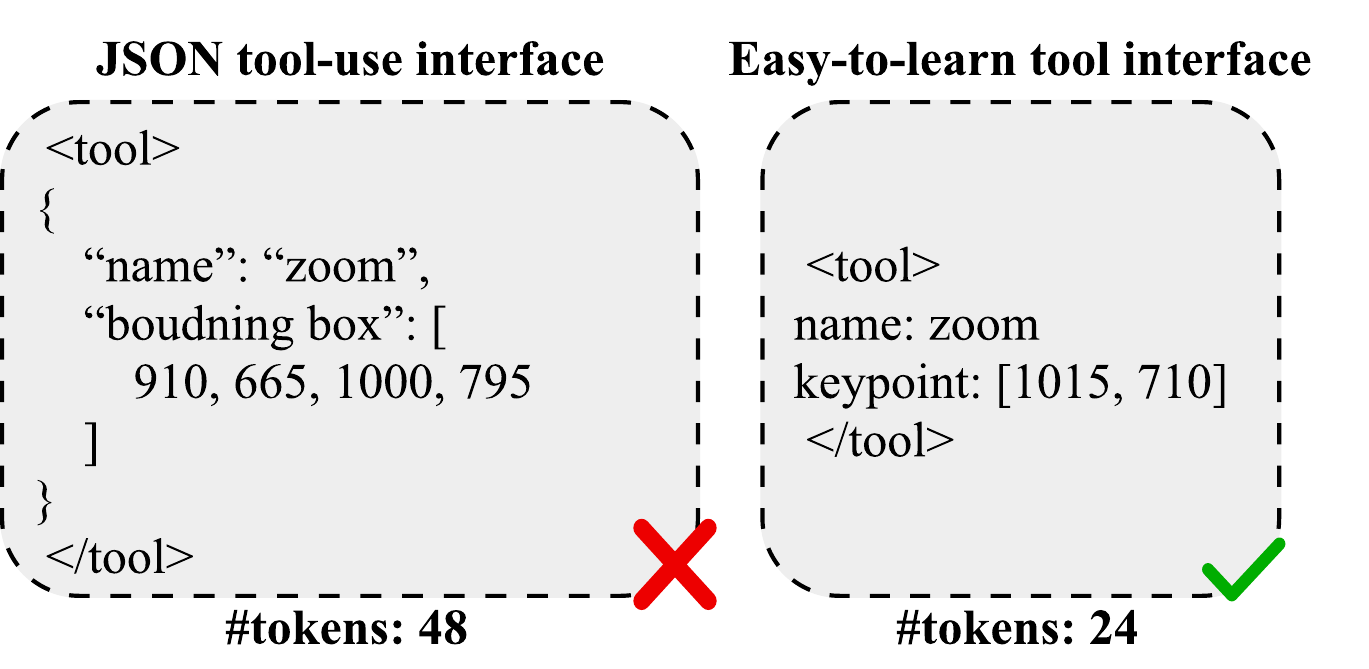}
  \end{center}
  \vspace{-1em}
  \caption{Our YAML-like interface for tool-use calls uses fewer tokens than standard JSON, and is easier for small models to format correctly.}
  \label{fig:tool-use-format}
\end{figure}
We find that under limited resource constraints, an easy-to-learn tool interface is essential. Our preliminary experiments were performed with a tool invoked with a complex format (i.e. JSON) accepting a bounding box in pixel coordinates as a parameter. However, training with this setup was frequently unstable. The model often struggled to generate correctly formatted tool calls, often hallucinating arguments or generating invalid JSON. We found that the model would gradually forget how to use the tool altogether. This led to complete model collapse in later training stages. Moreover, we found that this approach did not produce any improvements over the baseline model with no tools. The model learned to zoom into very small and often meaningless regions of the image. We hypothesized that a simpler format would be easier to learn. We achieved this in two ways. First, we adopted a YAML-like tool use interface that is less brittle and more concise than JSON (Figure~\ref{fig:tool-use-format}). This format contains fewer tokens, helping the model learn tool use more efficiently. Additionally, we changed the tool to accept a keypoint defined in pixel coordinates instead of a bounding box. The tool then returns a fixed 400 x 400 pixel crop centered at the keypoint. This allows for the model to be less precise in its localization and increases the likelihood that the tool's output contains information that is useful to the query. 

\subsection{Training}
Targeting model efficiency, we train a small Qwen2.5-VL-3B-Instruct model with GRPO on a subset of the training data from the TextVQA~\cite{singh2019towards} dataset. The TextVQA dataset evaluates a model's capacity to read, interpret, and reason about textual content in images to answer related questions. This dataset was chosen as representative of tasks that require focus on a particular detail in order to find the answer: in this case, to read text in a specific location in the image. Note that the model is not exposed to any high-resolution images during the training stage, but our approach helps the model generalize well on such visual tasks at inference time. The system prompt gives the model the option of calling a zoom tool anchored at a point of interest in the image with a fixed bounding box. One can inspect our system and user prompts, in addition to how we describe the tool to the model in Appendix \ref{sec:appendix.prompts}.

One critical implementation detail is ensuring the model is only trained on its own generated tokens, and not image tokens generated by the tool. To accomplish this, we mask out tokens generated by the tool during the GRPO loss computation. This prevents the model from receiving confusing gradients from tool outputs it never produced, which is essential for learning effective tool usage patterns.

\paragraph{Reward shaping}
The rewards for a successful visual reasoning trace are sparse, as the answer often follows from multiple reasoning steps involving an optional call to the external tool. To address this, we follow standard practice with GRPO and use structured rewards to guide learning. Specifically, we reward the model separately for answer correctness, for proper formatting, and for successful tool use. The total reward is computed as \( R = \alpha R_c + \beta R_f + \gamma R_t \), where the correctness reward \( R_c = \lambda R_a + (1 - \lambda) R_e \) combines a hard and a soft component. The hard reward \( R_a \) is the standard VQA score~\cite{singh2019towards}, computed as \( \min\left(\frac{\# \text{ matching human answers}}{3}, 1\right) \), which provides credit based on how many of the ten human annotators gave the same answer as the model on each example. The soft reward \( R_e \) is a partial-credit version of the VQA score that uses the average normalized edit distance between the model's answer and the three closest ground truth responses, providing a smoother signal on near-miss predictions even when the exact answer is not reproduced. $R_t$ provides reward on any successful use of the zoom tool, even if the selected region does not contain information relevant to the query. The weights $\alpha$, $\beta$, $\gamma$, and $\lambda$ are treated as tunable hyperparameters, chosen to balance the relative importance of each reward component.
  Our best-performing configuration used \( \alpha = 1 \), \( \beta = 1 \), \( \gamma = 0.1 \), and \( \lambda = 0.5 \).

\paragraph{Data mix}
We train our model on a curated mix of images from datasets where zooming in to see visual detail is beneficial. 
We find that the nature of the training data matters in obtaining the best results when it comes to teaching VLMs tool use. It is important to prioritize training on those images and tasks that are most difficult for the base model without access to our tool. We found the most success when the model was trained on the subset of the TextVQA training data for which the base Qwen2.5-VL-3B-Instruct model gets an average VQA score of $< 0.5$ from an 8-shot evaluation on the training set. This evaluation was carried out at a temperature of 1.0 and with at most 10 tokens generated per query. This filtering step reduced our training set size from approximately 34,600 examples to 7,800.

We originally explored training on the full TextVQA task, but found that tool use was not necessary for a large number of the images. Even when provided a small reward for correct tool usage, the model still learns to not use the zoom tool. We hypothesize that our base model was trained on this task already, although the exact data mix is not specified in \cite{bai2025qwen2}. Additionally, we observe that the model could directly read the text from the full image at the downsized resolution. Based on these results, we attempted to train on only the most difficult images in the training set, where the base model gets an average VQA score of $0$ from an 8-shot evaluation. This reduced the training set size to approximately 4,000 examples. This results in effective learning to use zoom, even in the absence of extrinsic tool use reward, but suffers from poor generalization. These results drove our final mix described above -- we over represent difficult examples while maintaining some easier ones representative of the full task distribution.





%% file: sections/experiments.tex
\section{Experiments}

\input{tables/main_results}

\subsection{Experimental Setup}
\paragraph{Evaluation benchmarks} We evaluate our models against baselines on both in-domain and out-of-domain high resolution benchmarks. The in-domain reasoning capabilities are evaluated on the validation set of TextVQA. Following existing works, we use VQA score to evaluate the models' performance on TextVQA~\cite{singh2019towards}. The out-of-domain benchmarks are $V^*$ Bench~\cite{Wu_2024_CVPR} and HR-bench~\cite{Wang_Ding_Zeng_Zhou_Shen_Luo_Yu_Tao_2025}, with fine-grained multi-choice question answering tasks used to assess the models' generalization capabilities. Both datasets evaluate a model's ability to answer questions about localized and fine grained details in high resolution images. $V^*$ Bench's images are an average image resolution of 2246 x 1582  pixels, while HR-bench provides 4k and 8k resolution variants. We evaluate on each example once with a temperature of 0.1 and a maximum of 2048 tokens generated per completion in a rollout.

\paragraph{Baselines} We compare our model to three categories of VLMs: 1) Proprietary VLMs, including Qwen-VL-Max~\cite{bai2023qwen} and GPT4-o~\cite{hurst2024gpt}; 2) visual search methods that engineered for high-resolution tasks, namely SEAL~\cite{Wu_2024_CVPR}, $\text{DC}^2$~\cite{Wang_Ding_Zeng_Zhou_Shen_Luo_Yu_Tao_2025}, and ZoomEye~\cite{shen2024zoomeye}; 3) open-source VLMs, such as LLaVA-HR-X-7B~\cite{zhang2024llava} and Qwen2.5-VL-Instruct~\cite{bai2025qwen2} series models.

\paragraph{Training details} We run our training on 4 A100 80GB GPUs executing the GRPO algorithm. To control the training memory footprint, we limit the rollouts to 3 and the maximum tool calls to 1, while the global batch size is set to 4. We train for 800 steps with a maximum of 2 images per rollout, a completion length of 2048 tokens per response in a rollout, and a temperature of 1.0. We use Adam with $\beta_2=0.98$, a cosine learning rate schedule with learning rate at $1e-6$ with 10 warm-up steps, a maximum grad norm of $1.0$, no KL regularization and a clip-higher strategy~\citep{yu2025dapoopensourcellmreinforcement} with $\epsilon_{low}=0.2$ and $\epsilon_{high}=0.28$. Our hyperparameter choices were informed by prior experience with training with GRPO on Qwen2.5-VL-3B and recent works \cite{zhan2025vision} and were further refined empirically as we developed our method. We present the results from the training run that performed the best and are representative of the approach’s effectiveness. Our training infrastructure builds on HuggingFace's TRL framework~\citep{vonwerra2022trl} and Verifiers~\citep{brown_verifiers_2025}, which we extended to support multi-turn rollouts, vision-language models, and structured tool use.

\subsection{Results}
\paragraph{RL substantially improves VLMs' performance on some high-resolution tasks} As shown in Table~\ref{tab:main-result}, compared to the Qwen2.5-VL-3B baseline, our model achieves 5.7\% overall accuracy improvement on $V*$Bench. At the same time, the 3B model trained with GRPO achieves similar performance to the 7B base model, demonstrating greater parameter efficiency given the comparable end-task accuracy. However, our approach decreases in-domain performance on the validation set of TextVQA and does not improve results on HR-Bench. We will discuss these limitations below. 

\paragraph{End-to-end RL is parameter-efficient compared to visual search methods}  
Compared to heavily engineered visual search approaches such as SEAL, our GRPO-trained VLM achieves comparable end-task performance on $V^*$Bench while being significantly more efficient. SEAL uses a 7 billion parameter model, whereas our model is only 3 billion parameters, yielding a smaller memory and computational footprint. Additionally, SEAL performs a hierarchical search and requires evaluating an average of 4.65 cropped image regions per query at inference time~\cite{Wang_Ding_Zeng_Zhou_Shen_Luo_Yu_Tao_2025}, resulting in substantial overhead. In contrast, our model performs a single targeted zoom-in action per query, yielding a simpler and faster inference process. Our tool-use formulation enables the model to efficiently select the most relevant region of interest without needing an explicit search policy at inference.

\paragraph{Low-resolution training does not generalize to 4K \& 8K images at inference time} Results in Table~\ref{tab:main-result} show that the performance of the model trained after GRPO is indifferent to the base model, suggesting that such performance gains shown on $V^*$Bench are not achieved on higher-resolution tasks, namely the HR-bench 4K and 8K tasks. We conjecture that the resolution gap between training and inference causes RL to generalize poorly, and our future work will focus on curating training data that supports better performance in these settings. We believe this will require incorporating higher-resolution images during training to bridge the resolution gap.

\subsection{\label{analysis}Discussion and analysis}
We analyze our results in more depth and discuss the contributions of design choices in this section. 

\begin{figure}
    \centering
        \includegraphics[width=\linewidth]{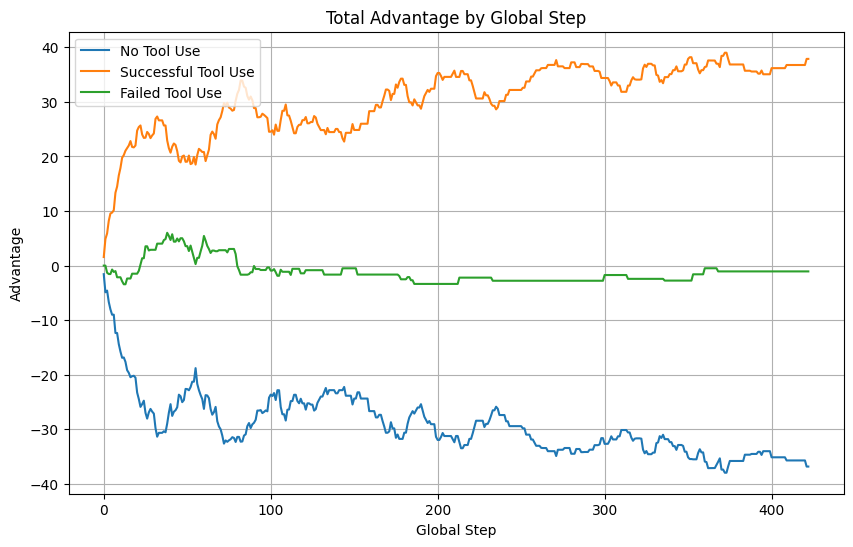}
        \caption{Accumulated GRPO advantage for policy trajectories with successful tool use (orange), failed tool use (green) or no tool use (blue) during a training run. }
        \label{fig:total-advantage}
\end{figure}

\paragraph{Effect of structured reward} 
Figure~\ref{fig:total-advantage} shows how the structured reward creates an advantage for successful tool use during training. We plot the total observed advantage grouped by the category of rollout: 1) no tool use; 2) successful tool use; 3) failed tool use. Our training paradigm ensures that successful tool-use calls will be reliably rewarded as training progresses. Moreover, we find that responses without any tool use will receive a negative advantage compared to failed tool-use calls. This supports our belief that our reward structure contains a good signal for the model to tackle tasks that require understanding of visual details in high-resolution images.

\paragraph{Effect of data mix choice}
Before using the training examples where the base model achieves $<$ 0.5 VQA score from 8-shot evaluation, we also explored training the model with difficult data only, where only the examples for which the base model gets an average of 0 VQA score are selected, as detailed in Section \ref{method:details}. We find that this data selection strategy hurts rather than helping the model's performance, resulting in 71.7\% accuracy on $V^*$Bench, which is lower than the baseline. We conjecture that this training distribution was not representative of the broader task distribution seen during evaluation. As a result, the model failed to generalize to the questions encountered at test time. Interestingly, this was the only setting in which the model learned to use the tool without any explicit reward for doing so ($\gamma = 0$), suggesting that tool use can emerge naturally when the model is sufficiently uncertain.

\begin{figure}
    \centering
        \includegraphics[width=0.9\linewidth]{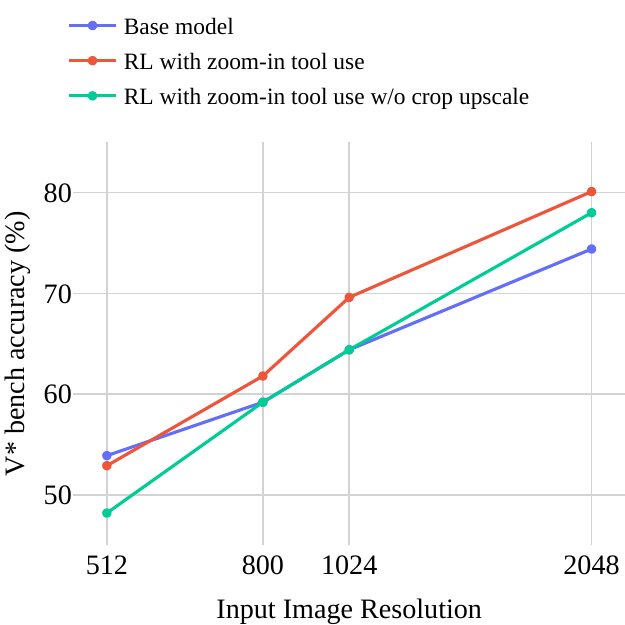}
        \caption{Overall accuracy of VLMs on $V^*$Bench with varied input image resolutions at inference time.}
        \label{fig:ablation-study}
\end{figure}

\paragraph{Effect of inference time image resolution}
As shown in Figure~\ref{fig:ablation-study}, we notice that using RL to train the VLM for tool-use substantially improves the model's performance on $V^*$Bench, even if image resolution is reduced at inference time. This implies that a potential application of our model is to perform low-resolution inference and use a zoom-in tool on high-resolution tasks, thereby improving the system's inference efficiency. Furthermore, we also notice that resizing the crop dimension to match the input image plays an essential role in accuracy improvements. Without applying this technique, the model's performance degrades to a baseline level.

%% file: tables/main_results.tex
\begin{table*}[htbp]
\centering
\small
\begin{tabular}{@{}l|ccc|ccc|ccc|c@{}}
\toprule
\textbf{Model}                                                 & \multicolumn{3}{c|}{$V^*$ Bench}                    & \multicolumn{3}{c|}{HR-Bench 4K}               & \multicolumn{3}{c|}{HR-Bench 8K}               & \multirow{2}{*}{$\text{TextVQA}_\text{val}$} \\ \cmidrule(r){1-10}
\textbf{}                                                      & \textbf{Attr.} & \textbf{Spatial} & \textbf{Overall} & \textbf{FSP} & \textbf{FCP} & \textbf{Overall} & \textbf{FSP} & \textbf{FCP} & \textbf{Overall} &                                              \\ \midrule
\textbf{Closed-source VLMs}                                    &               &                  &                  &              &              &                  &              &              &                  &                                              \\
Qwen-VL-max~\citep{bai2023qwen}                                & -             & -                & -                & 65.0         & 52.0         & 58.5             & 54.0         & 51.0         & 52.5             & 61.5                                         \\
GPT4-o~\citep{hurst2024gpt}                                    & -             & -                & 66.0             & 70.0         & 48.0         & 59.0             & 62.0         & 49.0         & 55.5             & 77.4                                         \\ \midrule
\textbf{Visual Search}                                         &               &                  &                  &              &              &                  &              &              &                  &                                              \\
SEAL~\citep{Wu_2024_CVPR}                                      & 74.8          & 76.3             & 75.4             & -            & -            & -                & -            & -            & -                & -                                            \\
$\text{DC}^2$~\citep{Wang_Ding_Zeng_Zhou_Shen_Luo_Yu_Tao_2025} & -             & -                & -                & 53.0         & 47.0         & 50.0             & 40.5         & 45.0         & 42.3             & -                                            \\
ZoomEye~\citep{shen2024zoomeye}                                & 93.9          & 85.5             & 90.6             & 84.3         & 55.0         & 69.6             & 88.5         & 50.0         & 69.3             & -                                            \\ \midrule
\textbf{Open-source VLMs}                                      &               &                  &                  &              &              &                  &              &              &                  &                                              \\
LLaVA-HR-X-7B~\citep{luo2024feast}                             & 51.3          & 64.5             & 56.5             & 57.8         & 46.3         & 52.0             & 42.0         & 41.3         & 41.6             & 67.1                                         \\
Qwen2.5-VL-7B~\citep{bai2025qwen2}                             & 80.9          & 76.3             & 79.1             & 85.2         & 52.2         & 68.8             & 78.8         & 51.8         & 65.3             & 84.9                                         \\
Qwen2.5-VL-3B~\citep{bai2025qwen2}                             & 81.3          & 63.2             & 74.4             & 81.8         & 48.5         & 65.1             & 80.5         & 47.3         & 63.8             & 79.3                                         \\ \midrule
Ours                                                           & 82.4          & 76.3             & 80.1             & 81.8         & 48.8         & 65.3             & 69.5         & 46.8         & 58.1             & 73.4                                         \\
\multicolumn{1}{r|}{$\Delta$ v.s. Qwen2.5-VL-3B}               & +1.1          & +13.1            & +5.7             & +0.0         & +0.3         & +0.2             & -11.0        & -0.5         & -5.7             & -5.9                                         \\ \bottomrule
\end{tabular}
\caption{Results on evaluation benchmarks. We examine the multi-choice QA accuracy of our method and baselines on $V^*$ Bench and HR-Bench. Attr. and Spatial represent the subtask of attribute recognition and spatial relationship reasoning in $V^*$ Bench, respectively. FSP and FCP are the Fine-grained Single-instance Perception and Fine-grained Cross-instance Perception subtasks in HR-Bench. We report VQA score when evaluating all models' performance on $\text{TextVQA}_\text{val}$.}
\label{tab:main-result}
\end{table*}

%% file: sections/related_work.tex
\section{Related Work}

\subsection{Reasoning with VLMs}
As VLMs naturally inherit the reasoning capabilities from their LLM backbone, researchers attempt to teach VLMs with chain-of-thought reasoning traces to ask the model to think before answer~\citep{luan2024textcot,bai2025qwen2}. Since the breakthrough of DeepSeek-R1~\citep{shao2024deepseekmath,guo2025deepseek}, researchers have been focusing on bringing the success of R1-like GRPO training paradigms to VLMs to improve models' visual reasoning performance. Recent research found that VLMs can achieve superior reasoning capabilities in vision-intensive mathematical tasks through GRPO~\citep{huang2025vision,zhan2025vision,shen2025vlm,yang2025r1,deng2025openvlthinker,wang2025vl}. However, existing work often requires expensive human annotation or distillation from larger models for generalization. In the meantime, limited effort has been spent on improving VLMs' vision-oriented reasoning capabilities, where the model can leverage task-specific details residing in images.







\subsection{Fine-grained Processing of VLMs}
Recent research has pointed out that VLMs often neglect detailed information in images~\citep{gou2024well}, and have blind faith in textual modality~\citep{Rahmanzadehgervi_2024_ACCV,deng2025words}. To address this issue, inspired by humans' behavior in recognizing information from images~\citep{Yang_2020_CVPR}, a series of visual search-based improvements on VLMs have been proposed. Even though such methods can inject additional information of salient regions in images to the VLM, they rely on extrinsic visual search components with complex searching algorithms~\citep{Wu_2024_CVPR,shen2024zoomeye,li2025dyfo}, thus limiting the efficiency of the system. Meanwhile, another line of research attempts to convert the input image into multi-resolution crops and feed them to the VLM end-to-end~\citep{thapa2024dragonfly,liu2024infimm,zhang2024llava,Wang_Ding_Zeng_Zhou_Shen_Luo_Yu_Tao_2025}. Although these methods successfully provide the VLM with image details to tackle downstream tasks, there is computational overhead introduced by the image tokens of the cropped patches. At the same time, these solutions crop the original image generally without recognizing the importance of specific regions in the image to tackle downstream tasks, which may limit their ability to help VLMs solve real-world problems.











\subsection{Multimodal Tool use with LLMs}
Researchers have been attempting to enable language models to do multimodal tasks by designing language model tool-use systems, i.e., developing systems that treat language models as black boxes to call APIs or use various tools~\citep{li-etal-2023-api,9851934}. Without training multimodal foundational models from scratch, previous research has found that iteratively prompting LLMs with vision foundation models can successfully perform numerous vision-language tasks~\citep{yang2023mm,wu2023visual,Gupta_2023_CVPR}. However, there is only limited research into applying the same approach to vision language models to enhance their vision-oriented reasoning capabilities~\citep{10943671}. 
A number of researchers have recently applied reinforcement learning to LLMs, including with successful tool-use learning~\citep{jin2025search,qian2025toolrl}. These efforts have been primarily limited to text-only models, leaving much to be explored in the multimodal domain. 

More recently, DeepEyes~\citep{zheng2025deepeyes}, corroborates some of our findings on using RL to encourage tool usage for vision. However, our work is distinct in its focus on identifying practical approaches and best practices for tool use under constrained computational resources, targeting smaller models and efficient training strategies.







%% file: appendix/prompt.tex
\onecolumn
\section{Prompts}
\label{sec:appendix.prompts}
Below, you can find: 
\begin{enumerate}
    \item The system prompt, which describes the desired response format and how tools can be invoked.
    \item The zoom tool description, which provides the model with the appropriate context to use the tool.
    \item The user prompt, which shows how we prompted the model to answer each question during training and evaluation.
\end{enumerate}

\small
\begin{tcolorbox}[title=System Prompt, colback=gray!5!white, colframe=black!75!black, fontupper=\ttfamily, sharp corners]
You may call any of the tools exactly one time. You have access to the
following tools to help solve problems: \\

\{tool\_descriptions\}\\

For each step:\\
1. Start by thinking through your reasoning inside <think> tags. Then
either return your answer inside <answer> tags, or use a tool inside
<tool> tags.\\

2. If needed, use a tool by writing its arguments inside <tool> tags.
Use one line for each argument in the format "key: value". The first
line must be "name: <tool\_name>".\\

3. You will see the tool's output inside <result> tags.\\

4. Continue until you can give the final answer inside <answer> tags.\\

Tools expect specific arguments. Follow the examples carefully for the
required keys and expected value formats. \\

Do not make up tools or arguments that aren't listed.
\end{tcolorbox}

\begin{tcolorbox}[title=Zoom Tool Description, colback=gray!5!white, colframe=black!75!black, fontupper=\ttfamily, sharp corners]

Zoom Tool: Returns the image zoomed in on the specified keypoint. \\

This is useful to see a region of interest with higher clarity, like for reading text or identifying objects. \\

Args: \\
    \hspace*{2em}keypoint: list[int], the [x, y] coordinates of the point to center the zoom on. \\
    \hspace*{2em}Must be within the image boundaries. \\

Returns:\\
    \hspace*{2em}The image zoomed in on the specified keypoint. \\

Examples:\\
    \hspace*{2em}<tool> \\
    \hspace*{2em}name: zoom \\
    \hspace*{2em}keypoint: [500, 400] \\
    \hspace*{2em}</tool> \\
    
    \hspace*{2em}<tool> \\
    \hspace*{2em}name: zoom \\ 
    \hspace*{2em}keypoint: [23, 44] \\
    \hspace*{2em}</tool> \\
    
    \hspace*{2em}<tool> \\
    \hspace*{2em}name: zoom \\
    \hspace*{2em}keypoint: [723, 461] \\
    \hspace*{2em}</tool> \\

\end{tcolorbox}

\begin{tcolorbox}[title=User Prompt, colback=gray!5!white, colframe=black!75!black, fontupper=\ttfamily, sharp corners]

The image size is \{image\_size\}.\\
Please thoroughly think through the question and refine your answer while thinking. You should try to collect the visual evidence you need to support your answer. Then, provide your answer. The answer (which you will provide in the <answer> </answer> tags) should be a single word or phrase directly answering the question.\\
Question: \{question\}
\end{tcolorbox}
\twocolumn